# Indoor Testing and Simulation Platform for Close-distance Visual Inspection of Complex Structures using Micro Quadrotor UAV

Zhexiong Shang[1] and Zhigang Shen[2]

1) Ph.D. Candidate, Durham School of Architectural Engineering & Construction, University of Nebraska-Lincoln, Lincoln, Nebraska, USA. Email: szx0112@huskers.unl.edu
2) Ph.D., Associate Professor, Durham School of Architectural Engineering & Construction, University of Nebraska-Lincoln, Lincoln, Nebraska, USA. Email: shen@unl.edu

**Abstract:**

In recent years**,** using drone, also known as unmanned aerial vehicle (UAV), in close-distance visual inspection has became an active area in many disciplines. However, many challenges still remain before we can achieve autonomous inspection, especially when inspecting complex structures. The complex civil structures, such as bridges, dams and wind turbines, are large-scale and geometrical complicated. It requires sophisticated path planning algorithms to achieve close-distance inspection and, at the same time, avoid collisions. In practice, directly deploying the path planning result on such structures is error prone, costly, and full of hazards. In this paper, rely on micro quadrotor UAV, the authors present an affordable experimental platform for testing drone-based path planning result. The platform allows the users to conduct many path planning experiments at any time without worrying expensive and time consuming outdoor test flying. This platform is developed based on the bundle of Crazyflie, which includes Crazyflie 2.0 quadrotor, Crazyradio and loco positioning system (LPS). Equipped with an onboard micro FPV camera, the visual data can be lively streamed to the host computer during flight. The functions of manual configuration and waypoints control are explicitly designed in this platform to increase its flexibility and performance on path following and debugging. To evaluate the practicability of the proposed test platform, two existing drone-based path planning algorithms are tested. The results show that even though certain level of error existed, the quality of visual data and accuracy of path following are high enough for simulating most practical inspection applications.

## 1. INTRODUCTION

Inspection of complex civil infrastructures, such as bridges, dams and wind turbines, is a particular important application in civil and construction industry. The traditional methods rely on site inspector and ground equipment that are not only costly, but also full of hazards as fall from height and struck-by are two leading factors of fatalities in the field of construction (OSHA, 2016). The ongoing developments of drones, also known as unmanned aerial vehicle (UAV), has led to a growth of aerial inspection in various disciplines (Baiocchi, Dominici, & Mormile, 2013; Rau, Jhan, Lo, & Lin, 2011; Samad, Kamarulzaman, Hamdani, Mastor, & Hashim, 2013; Zhang et al., 2012). UAVs are quicker, safer and cost effective. It can easily reach areas that are inaccessible to ground vehicles and undertake tasks that are dangerous to human.

Currently, the primary sensor used on UAV are optical sensors because of its low cost, light weight and easy to implement. The existing studies showed that with onboard camera, UAV can achieve at a high level of accuracy even for professional applications (Metni & Hamel, 2007). In construction industry, research on visual inspection using UAV is still at a preliminary stage, the current applications include construction progress monitoring (Lin, Han, & Golparvar-Fard, 2015; Siebert & Teizer, 2014), 3D site reconstruction (Chen & Cho, 2016; Shang & Shen, 2018) and structural inspection (Bulgakow, Bock, & Sayfeddine, 2014; Ellenberg, Branco, Krick, Bartoli, & Kontsos, 2014; Khan et al., 2015; Reagan, Sabato, Niezrecki, Yu, & Wilson, 2016). However, in most cases, UAVs are still manually controlled by a remote operator. Thus, its application performance is highly dependent on the experience and skills of the operator. Due to the rising complexity of civil structures and its clutter environment, the human errors can easily cause unnecessary flight time or even avoidable crashes.

Thus, automatous flight is needed to facilitate site inspection and improve the safety redundancy. In the field of robotics, autonomous inspection using ground, maritime or aerial robots has recently been applied in various applications such as submarine structure inspection (Englot & Hover, 2010), wind turbine damage search (Schäfer, Picchi, Engelhardt, & Abel, 2016) and urban surveillance (Pettersson & Doherty, 2004). These methods developed sophisticated algorithms to generate paths that can achieve close-distance and high quality inspection. However, up to now, only few studies validate such algorithms on field tests. In many cases, the aerial inspection was restricted by the local law that special permission is required for UAV to fly at public area. For large-scale and complex civil infrastructures, directly deploying the flight test is error prone, costly, needs specific traffic control and rely on the weather conditions.

In last years, micro UAVs are widely used by academics in various applications. Comparing to the medium and large level drones, micro UAV has the advantages of low cost of maintenance and has nearly no safety concern to its surroundings, which make it a perfect tool to testing new algorithms. Current applications of micro UAVs have been grown to many areas, such as control system design (Mahony, Kumar, & Corke, 2012), path finding (Giernacki, Skwierczyński, Witwicki, Wroński, & Kozierski, 2017) and swarms (Preiss, Honig, Sukhatme, & Ayanian, 2017). However, until now, its applications on testing the inspection path planning algorithms of complex structures is still missing.

Thus, to diminish the unnecessary cost and increase the efficiency of UAV inspection path planning progress, in this study, the authors present an affordable experimental platform that allows researchers to physically and digitally test their autonomous inspection ideas in an indoor environment before the expensive and time consuming outdoor tasks. Crazyflie 2.0 ("https://www.bitcraze.io,"), which is a fully open-source micro quadrotor, is selected as the test UAV in this platform. Taken a planned path as input, the platform sequentially transfers the path into waypoints and communicate with Crazyflie drivers to achieve autonomous indoor flight test. A FPV camera is mounted at the top of Crazyflie such that the visual data can be both broadcasted in the air and assessed after flight. To validate the practical value of the proposed platform, the initial results of visual representation and trajectory following of two planned paths on an arbitrary inspection target are presented.

## 2. ARCHITECTURE OF THE TEST PLATFORM

The general architecture of the proposed test platform is shown in Figure 1. The platform is developed based on the bundle of Crazyflie, which includes the Crazyflie 2.0, Crazyradio and bitcraze loco positing system (LPS). Crazyflie has two microcontrollers: the main microcontroller and the second microcontroller. The main microcontroller (STM32) runs FreeRTOS as the operating system to execute state estimation and low level attitude control through onboard sensors. The second microcontroller (nRF51) is used for wireless communication and power management. Communicated with the Crazyradio, a 2.4 GHz USB dongle, Crazyflie can be remotely controlled by a PC desktop/laptop computer with Linux/Window system. LPS is an indoor positioning system that can provide absolute 3D positions of Crazyflie with accuracy up to 10 cm. The system relies on Ultra Wideband (UWB) to measure the distances between a deck attached on Crazyflie and several fixed anchors in the space, and estimate the real-time position of the deck with Two Way Ranging (TWR) protocol and extended Kalman filter (EKF). Although the accuracy of LPS is not as high as many commercial local positioning systems, the trade-off is the cost efficiency. An Xbox controller is connected to desktop to enable real-time human interaction of autonomous inspection, and a FPV camera is mounted at the top of Crazyflie as the onboard visual sensor. The selected micro FPV camera is Goqotomo GT02 with clover antenna. It can send 40 radio channels on the band of 5.8 GHz. The camera is soldered at the top of the drone's battery holder such that the power support for Crazyflie can be shared. Connected with a video receiver, the real-time visual data are lively streamed to the desktop/laptop's monitor. These attached hardware makes the total weights of Crazyflie to 37 grams but still within the range of its maximum take-off limits of 42 grams.

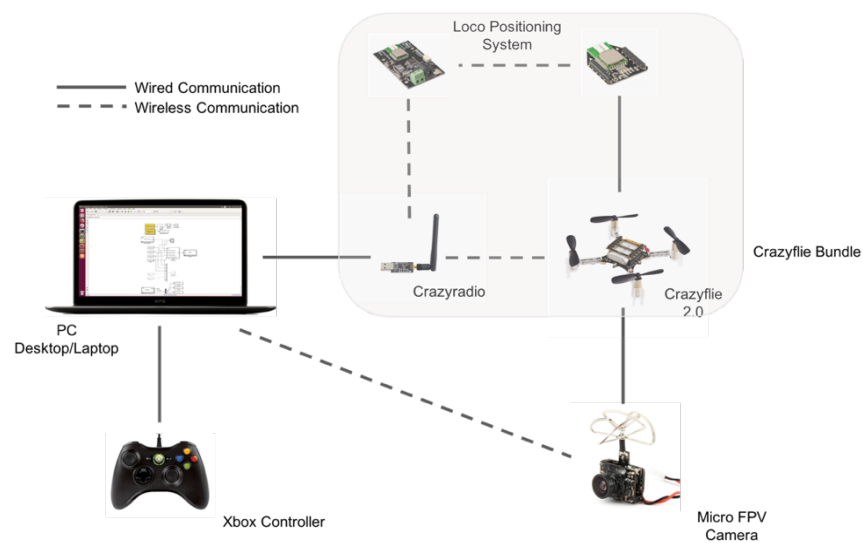

Figure 1. Architecture of the test platform

## 3. WORKFLOW OF THE TEST PLATFORM

Figure 2 shows the general workflow of the proposed platform. In this study, the authors assume an inspection path is given, designing specific path planning algorithms is out the scope of this research. The path is initially passed into Matlab Simulink which organizes the dataset as an array of 4D waypoints (x, y, z, yaw). The waypoint sequence is then transferred into the Message type of ROS with a pre-defined transmission rate. In this platform, ROS is not only applied as the interface between Simulink and Crazyflie driver, but also provides all the functionality of autonomous flight. In order to increase the performance of autonomous visual inspection, in this study, two functions are explicitly designed, which are: (1) *manual configuration*; and (2) *waypoint control*. *Manual configuration* allows users to manually send commands to the drone to initiate/pause/stop the autonomous flight. These commands are sent by hacking a Xbox controller (Figure 3 (left)). The autonomous flight will be only initiated when the drone is taken off and the mode is switched to *autonomous*. This setup not only minimize the potential crashes, but also give the flexibility for human interaction during the flight such that the inspection progress can be controlled and evaluated in a timely manner.

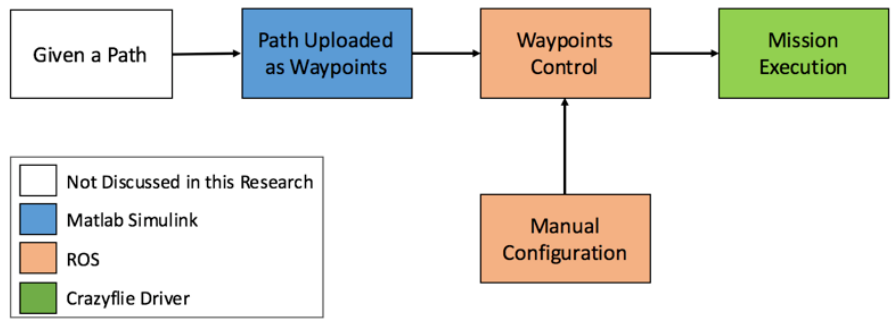

Figure 2. Workflow of the test platform

Another function is *waypoint control*. It controls when next waypoint should be sent to the micro UAV. To achieve this goal, a feedback waypoint control loop is designed. It sequentially checks whether the drone is arrived at the current waypoint by measuring if the drone's center of gravity is within a bounding box (shown in Figure 3 (right)). The bounding box is a virtual cube with center located at each designed waypoint location and side length equal to 15 cm. The side length is determined by the referenced precision of LPS. Only when the drone stays in the box for more than 0.5 second, the system will confirm the drone's arrival, and then the new waypoint will be updated. This process iterates until all designed waypoints are visited. To reduce the ground effect, the authors also increase the integral parameter of z direction to 40 in the PID position controller of Crazyflie Firmware to stabilize the UAV even when it flies close to the ground.

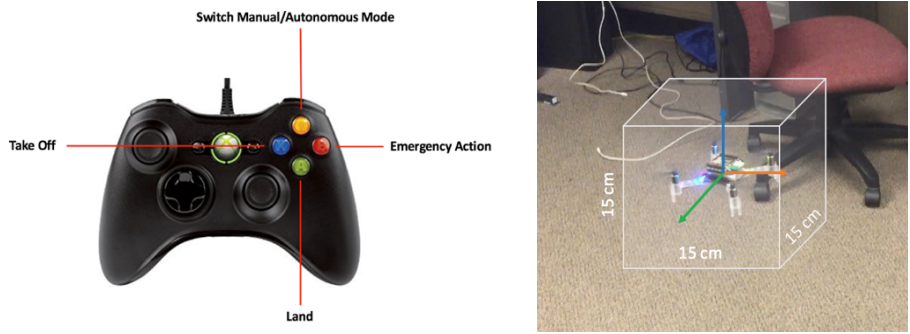

Figure 3. (left) Xbox controller with defined command in manual configuration; (right) Bounding box defined in waypoint control

## 4. APPLICATION EXAMPLE

To validate the practical value of the proposed platform, the indoor flight tests are presented. A set of boxes located at the center of the world coordinates of LPS is used as the target of inspection. Two path planning algorithms are applied in this example: (1) Spiral path and (2) random sampling path (Bircher et al., 2015). The spiral path is capable of inspecting many simple structures, such as chimney and wind turbines. It draws collision-free circles around the object with fixed intervals at vertical direction. The random sampling algorithm is a more complex inspection path planning algorithm. It rely on convex optimization to iteratively identify optimal viewpoints based on structure geometry and connects these viewpoints with a shortest path (Helsgaun, 2000). The pre-computed inspection paths for the inspection target using both algorithms are respectively shown in Figure 4 below. Although the sampling-based method is algorithmically more efficient than the spiral path, it does not consider the dynamic constraints of UAV. The sharp orientations change between adjacent waypoints give more challenges for UAV's path following in the real world implementations.

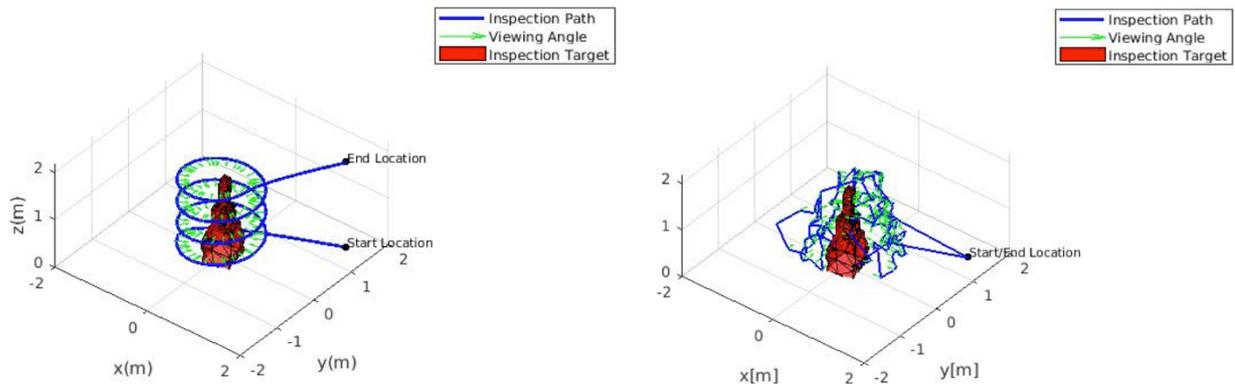

Figure 4. (left) Spiral path designed for the inspection target; (right) Random sampling path designed for the inspection target

To implement the planned paths into the test platform, the authors set the average traveling velocity of the drone as 0.3 meters per second. The velocity is tuned by considering both the accuracy of path following and the battery constraints of the FPV embedded Crazyflie. The onboard FPV camera is set with 30 fps and 600 VTL. The inspection scenario, FPV Crazyflie set up, and FPV image views during flight are shown in Figure 5.

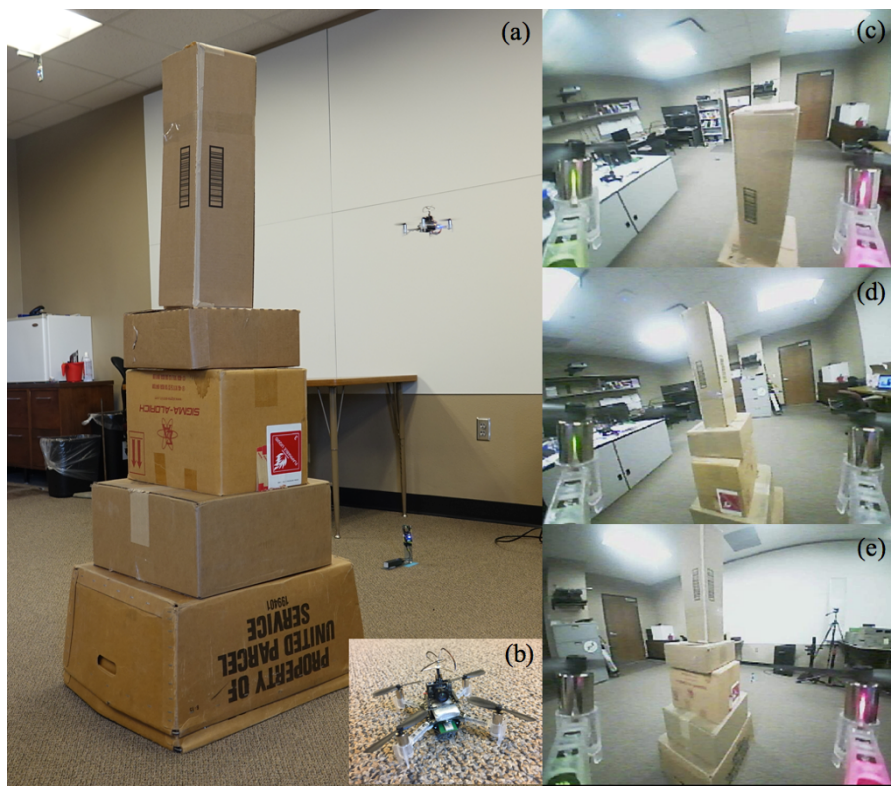

Figure 5. (a). Visual inspection scenario with inspection target and micro UAV in the world coordinate of LPS; (b). Detailed view of FPV Crazyflie setup; (c), (d), (e): Snapshots of visual images captured at different waypoints during the progress of the flight test.

Comparing the difference between the real trajectory and the planned path is critical to evaluate the accuracy and latency of the platform. The post-processed results of path following of both path planning algorithms at x/y/z dimensions are respectively plotted in Figure 6. The plots show that even though the average errors remain to be larger than 10 cm, the trajectories of both paths are highly uniform with the computed paths. It can be noticed that for both cases, the trajectories in z direction have a higher absolute error (17.25 cm) than in x/y dimensions (11.93 cm). Such difference can be caused by the vertical force induced by gravity and the errors of barometers.

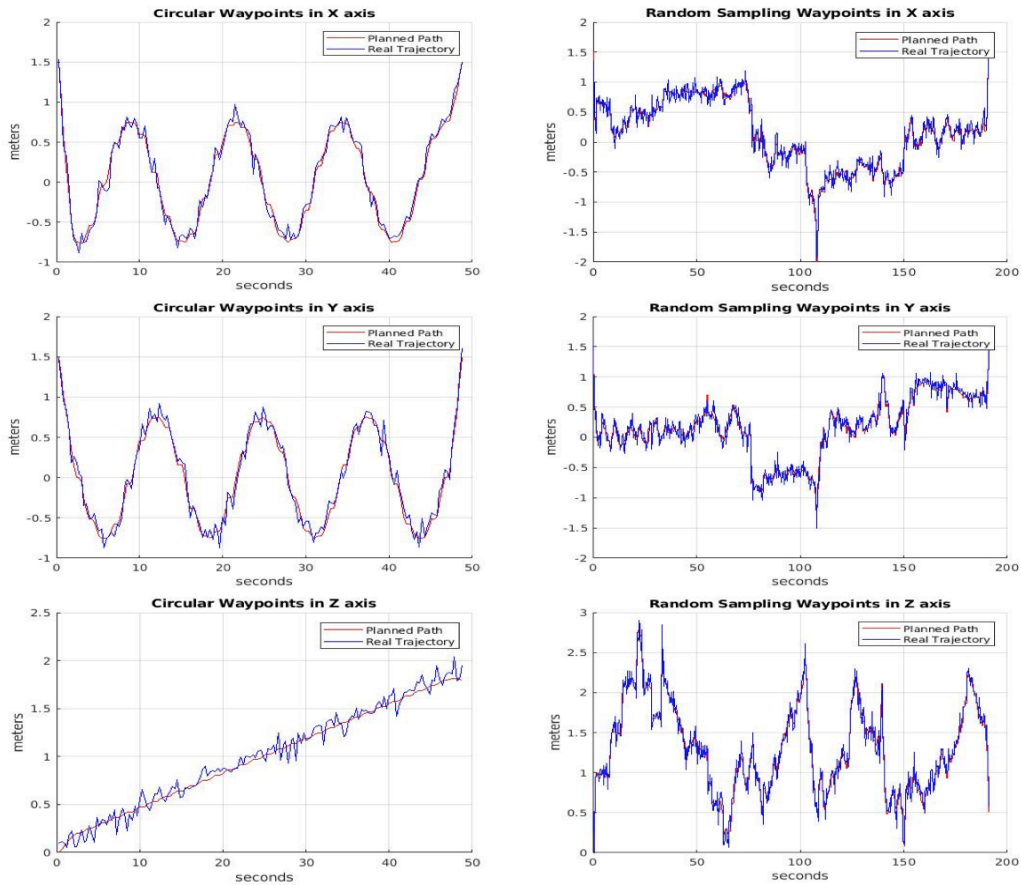

Figure 6. (left): The comparison of planned spiral path and real trajectory of UAV in x/y/z dimensions; (right): the comparison of planned random sampling path and real trajectory of UAV in x/y/z dimensions

## 5. CONCLUSION

In this study, the authors introduce a micro-UAV based in-door testing platform for autonomous visual inspection of complex structures. The proposed platform has many advantages over the direct field implementation which is both time consuming and costly. The platform is developed based on the original bundle of Crazyflie and LPS. To enable the visual inspection ability, a FPV camera is mounted at the top of UAV to broadcast visual images to the host computer in the air. To understand the practical value of the platform, a case study is presented to evaluate the quality of visual data and the accuracy of path followings results. The initial tests showed that the platform can provide valid visual data during inspection and the inspection trajectories are highly uniform with the computed path even though the paths are generated based on different algorithms. In the future studies, the authors will use this platform to test and evaluate different path planning results in order to achieve fully autonomous and high performance aerial inspection applications.